\documentclass{ecai}
\usepackage{times}
\usepackage{latexsym}
\usepackage{url,hyperref,lineno,microtype}
\usepackage{graphicx, color}
\usepackage{makecell}
\usepackage{amssymb, amsmath}
\usepackage{tabularx, booktabs, array}
\usepackage{xcolor}
\usepackage[linesnumbered, ruled, vlined]{algorithm2e}
\usepackage{multirow, float}
\usepackage{IEEEtrantools}

\ecaisubmission

\DeclareMathOperator{\softmax}{softmax}
\DeclareMathOperator{\abs}{abs}
\DeclareMathOperator{\concat}{concat}
\DeclareMathOperator{\floor}{floor}
\DeclareMathOperator{\LN}{LN}
\DeclareMathOperator{\mean}{mean}

\begin{document}

\title{Delving Deeper into the Decoder for Video Captioning}

\author{Haoran Chen
\and Jianmin Li
\and Xiaolin Hu \institute{
State Key Laboratory of Intelligent Technology and Systems, 
Institute for Artificial Intelligence, 
Beijing National Research Center for Information Science and Technology, 
THBI, 
Department of Computer Science and Technology, 
Tsinghua University, Beijing 100084, China., 
email: chr17@mails.tsinghua.edu.cn, \{lijianmin, xlhu\}@tsinghua.edu.cn, 
corresponding author: Xiaolin Hu.} 
}

\maketitle
\bibliographystyle{ecai}

\begin{abstract}
Video captioning is an advanced multi-modal task which aims to describe a video clip using a natural language sentence. 
The encoder-decoder framework is the most popular paradigm for this task in recent years. 
However, there exist some problems in the decoder of a video captioning model. 
We make a thorough investigation into the decoder and adopt three techniques to improve the performance of the model. 
First of all, a combination of variational dropout and layer normalization is embedded into a recurrent unit to alleviate the problem of overfitting. 
Secondly, a new online method is proposed to evaluate the performance of a model on a validation set so as to select the best checkpoint for testing. 
Finally, a new training strategy called \textit{professional learning} is proposed which uses the strengths of a captioning model and bypasses its weaknesses. 
It is demonstrated in the experiments on Microsoft Research Video Description Corpus (MSVD) and MSR-Video to Text (MSR-VTT) datasets that 
our model has achieved the best results evaluated by BLEU, CIDEr, METEOR and ROUGE-L metrics 
with significant gains of up to 18\% on MSVD and 3.5\% on MSR-VTT compared with the previous state-of-the-art models.
\end{abstract}

\section{Introduction}
The video captioning task aims to automatically generate a human-readable sentence to describe the content of a video clip that is usually 10 to 30 seconds long. 
Videos ``in the wild'' cover a variety of scenes and actions. 
Objects and relations between them in a video clip are required to be captured in order to determine nouns and space relations in a sentence. 
It is also necessary to model temporal relations in order to describe an event that lasts for a few seconds. 
It is therefore a difficult challenge to learn a model to generate an adequate description for a short video clip. 

The encoder-decoder framework has become the mainstream approach in the field of video captioning motivated by the success in machine translation and image captioning. 
For the decoder, it is most common to utilize long-short term memory (LSTM) \cite{DBLP:journals/neco/HochreiterS97} or gated recurrent unit (GRU) \cite{DBLP:conf/emnlp/ChoMGBBSB14} to map low-dimensional video representation, 
produced by the encoder, to a variable length sequence. 
Attention mechanism is also widely applied in video captioning models to generate visual features dynamically as input to recurrent neural network (RNN) units according to different contexts 
\cite{DBLP:conf/acl/PasunuruB17, DBLP:conf/emnlp/PasunuruB17, DBLP:journals/tmm/GaoGZXS17, DBLP:conf/iccv/HoriHLZHHMS17, DBLP:conf/cvpr/WangCWWW18}.

However, there are still some non-negligible problems in the decoder for video captioning. 
Firstly, video caption decoders usually suffer from the serious problem of overfitting. 
Dropout \cite{DBLP:journals/jmlr/SrivastavaHKSS14} is a common technique to prevent overfitting in convolutional network (ConvNet), 
but its application in recurrent networks is not so effective as researchers expected\cite{DBLP:journals/corr/ZarembaSV14}. 
Several methods have been proposed to apply dropout to recurrent networks \cite{DBLP:conf/asru/MoonCLS15, DBLP:conf/coling/SemeniutaSB16}. 
Variational dropout is a theoretically grounded method that has been adopted widely in kinds of deep learning frameworks \cite{DBLP:conf/nips/GalG16}. 
However, variational dropout slows down convergence speed and increases training time significantly. 
Layer normalization is proposed to accelerate convergence speed of RNN by stabilizing internal dynamics \cite{Ba2016LayerN}. 
As far as we know, the effect of layer normalization has not been explored in the context of video captioning before.

Then, it is a common practice that the loss value on the validation set is used as a metric to choose a model for testing. 
However, some metrics from natural language processing (NLP), 
e.g. BLEU\cite{DBLP:conf/acl/PapineniRWZ02}, METEOR\cite{DBLP:conf/acl/BanerjeeL05}, 
CIDEr\cite{DBLP:conf/cvpr/VedantamZP15} and ROUGE-L\cite{lin-2004-rouge}, 
are exploited to assess model's performance in testing. 
The divergence of evaluation between validation and testing phases leads to deteriorated performance in inference. 
Some researchers utilize one of those metrics to choose the best model for testing, such as BLEU and CIDEr. 
However, single metric can't reflect the overall performance of video captioning system since all of these metrics are reported in video captioning literature. 

Last but not least, most of the previous training algorithms have a common defect that 
they treat all the training samples equally which leads to what we call ``absolute equalitarianism'' in learning. 
The model trained in this way is likely to learn an intersection of the annotations for each video which consists of frequent words and phrases 
and is inclined to forget advanced words and complicated sentence structures since they vary too much from sentence to sentence. 

In this work, we propose three methods to solve each of these problems accordingly. 
First, variational dropout is used to reduce overfitting of the decoder and layer normalization is employed to counteract the prolonged training time brought on by dropout usage. 
Secondly, a new selecting method is proposed to choose the best model for testing based on comprehensive consideration of the various metrics. 
Besides, a novel training strategy, called \textit{professional learning}, is presented 
which trains the model in \textit{teacher forcing} way to learn basic knowledge using all the annotations equally and then optimize the same model with emphasis upon the samples it is good at. 
We perform extensive experiments on MSVD (YouTube2Text) \cite{DBLP:conf/iccv/GuadarramaKMVMDS13} and MSR-VTT \cite{DBLP:conf/cvpr/XuMYR16} datasets. 
And empirical results prove the effectiveness of the proposed methods in video captioning. 

\section{Related Work}
We pay our attention to regularization and training strategies for RNN as well as encoder-decoder-based video captioning literature in deep learning 
for the reason that they are highly correlated with our work. 

\subsection{Video Captioning}
Semantic information has been widely applied to assist video captioning models in generating annotations. 
Semantic SVO triplets and semantic hierarchies are exploited to output a brief sentence that summarizes the content in a video in \cite{DBLP:conf/iccv/GuadarramaKMVMDS13}. 
Tags of a video, i.e. key words from human annotations, are joined together with RNN parameters by matrix factorization technique to gain better sense of themes for images/videos \cite{DBLP:conf/cvpr/GanGHPTGCD17}. 
In another work, visual features and sentences embedding are projected to a joint low-dimension space. 
Semantic consistency between sentence content and video visual information is guaranteed by minimizing the distance between two embedded vectors in the joint space \cite{DBLP:journals/tmm/GaoGZXS17}.
Higher-order object interactions are modeled to improve the performance of video understanding \cite{DBLP:conf/cvpr/MaKMKAG18}.
High level semantic features derived from a video action classifier and an object detector is utilized to enrich video representation features in \cite{DBLP:conf/cvpr/AafaqALGM19}. 
We notice that classification results produced by video action classifiers and image classifiers have hardly utilized in previous works 
and we find that they can be naturally integrated into the semantic information provided by video tagging networks. 
With enhanced semantic information for video, a model is able to describe it in a more comprehensive way.

Inspired by successful application of attention mechanism in machine translation \cite{DBLP:journals/corr/BahdanauCB14}, object detection \cite{DBLP:journals/corr/BaMK14} and image captioning \cite{DBLP:conf/icml/XuBKCCSZB15}, 
attention mechanism has been applied to video captioning task in various ways\cite{DBLP:conf/iccv/YaoTCBPLC15, DBLP:journals/tmm/GaoGZXS17, DBLP:conf/iccv/HoriHLZHHMS17, 
DBLP:conf/acl/PasunuruB17, DBLP:conf/emnlp/PasunuruB17, DBLP:conf/cvpr/WangCWWW18, DBLP:conf/naacl/WangWW18, DBLP:journals/ijon/WuWCSSW18,DBLP:conf/cvpr/PeiZWKST19}. 
Attention mechanism contributes to the caption generation by distilling useful information dynamically according to the runtime context. 
Though it is extremely popular in recent years, we find attention mechanism does not always contribute to the performance of a model since the video captioning system may overfit on the training set more easily.

External or internal knowledge from a dataset is also utilized to provide more information for video captioning models. 
In MARN, memory structure is proposed to learn the relationship between a word and its various related visual features in order to achieve a more comprehensive understanding of the video content \cite{DBLP:conf/cvpr/PeiZWKST19}.
In TAMoE, external Wikipedia corpus is explored by primitive experts and 
it helps to transfer the knowledge learned from seen topics to unseen topics 
as well as improve the quality of generated captions on seen topics \cite{DBLP:conf/aaai/WangWZ0W19}. 

\subsection{Regularization and Training Strategies for Recurrent Neural Network}
Dropout and normalization operation are two common regularization methods for RNN. 
Dropout is a simple technique to reduce overfitting in neural networks but it does not work very well in RNN \cite{DBLP:journals/jmlr/SrivastavaHKSS14}. 
A new method of applying dropout shows that the dropout operator should only be applied to the non-recurrent connections \cite{DBLP:journals/corr/ZarembaSV14}. 
Dropout is proved to be a Bayesian approximation for representing uncertainty \cite{DBLP:conf/icml/GalG16} 
and variational dropout is proposed for RNN accordingly \cite{DBLP:conf/nips/GalG16}. 
Another method, that dropout mask should only be used on the update vector in RNN, is only supported by experiments \cite{DBLP:conf/coling/SemeniutaSB16}. 
It is a common phenomenon that dropout prolongs training time. 
Batch normalization is proposed to tackle the problem of internal covariate shift so that training process is accelerated \cite{DBLP:conf/icml/IoffeS15}. 
However, it requires different running averages of input statistics at different time steps when applied to an RNN unit which hinders its application in variable length sequence training. 
It has been supported by experiments that RNN, especially with long sequences, benefits significantly from Layer Normalization \cite{Ba2016LayerN}. 
 
The most common and intuitive training strategy is \textit{teacher forcing} which can be traced back to the end of 80s in 20th century\cite{DBLP:journals/neco/WilliamsZ89}. 
The ground truth $\boldsymbol{d}(t)$ for a sequence $\boldsymbol{s}$ is exploited as a part of the input $\boldsymbol{x}(t+1)$ to the recurrent unit during training. 
This leaves wide divergence between training process and inference phase since the unit output $\boldsymbol{y}(t)$ is utilized as the input $\boldsymbol{x}(t+1)$ to the recurrent unit. 
In another word, it is called exposure bias. 
A model trained by \textit{teacher forcing} is unlike to adapt to testing process because of the divergence.
A training method called scheduled sampling is proposed to minimize the gap between training phase and inference phase, 
which gently transfers the training phase from using ground truth as part of input for the recurrent unit to using model-generated tokens as part of input\cite{DBLP:conf/nips/BengioVJS15}. 
Although the problem of the divergence between training process and testing process is alleviated, the optimization goal of the scheduled sampling deviates from the natural-looking sentences \cite{DBLP:journals/corr/Huszar15}. 
Adversarial domain technique, called \textit{professor forcing}, is exploited to align dynamics of RNN during training and inference. 
\textit{Professor forcing} can, to some extent, act as a regularizer and has better ability to capture the long-term dependencies reported in \cite{DBLP:conf/nips/GoyalLZZCB16}.
Besides, reinforcement learning methods are proposed to train recurrent networks, 
e.g. self-critical sequence training (SCST) \cite{DBLP:conf/cvpr/RennieMMRG17}, 
CIDEnt-reward model \cite{DBLP:conf/emnlp/PasunuruB17}, Hierarchical Reinforcement Learning (HRL) \cite{DBLP:conf/cvpr/WangCWWW18,DBLP:conf/acl/WuRLS19}. 
Multi-task learning helps a model produce better input representation and improves generalization of a particular task by jointly training a model with related tasks \cite{DBLP:conf/acl/PasunuruB17}.
In curriculum learning, samples are introduced to train the model according to a predefined and fixed schedule called curriculum. 
All the samples are treated equally once introduced during optimization process \cite{Bengio:2009:CL:1553374.1553380}. 
 
Unfortunately, from the perspective of the whole training process, 
all of these training strategies treat training samples equally 
which leads the model to learn from a small intersection of annotations corresponding to each video. 
It results to limited vocabulary and repeated sentences in generated captions. 

\section{Model and Proposed Methods}
\label{section:methods}
The encoder-decoder framework \cite{DBLP:conf/emnlp/ChoMGBBSB14, venugopalan-etal-2015-translating} for video captioning is adopted in our work. 
In the encoder, a video $\boldsymbol{vid}$ is split into $K$ frames. 
A pre-trained image classifier is applied on each frame and it outputs feature maps $\boldsymbol{v}_{img, k}, k=1,2,...,K$ and classification results $\boldsymbol{cls}_{img, k}$ for each frame. 
Spatial feature map $\boldsymbol{v}_{img}$ and image classification result $\boldsymbol{cls}_{img}$ are obtained for each video by 
applying average operation over time axis on $\boldsymbol{v}_{img, k}$ and $\boldsymbol{cls}_{img,k}$. 
A pre-trained video action classifier is employed to produce global spatio-temporal feature map $\boldsymbol{v}_{vid}$ and action classification result $\boldsymbol{cls}_{vid}$ over all the frames. 
A tagging network is trained to generate semantic tags $\boldsymbol{s}_{t}$ for each video. 
On MSVD dataset, the output of the encoder for a video clip $\boldsymbol{vid}_i$ is composed of visual spatio-temporal features $\boldsymbol{v}_i$ 
and semantic information $\boldsymbol{s}_i$:
\begin{equation} \label{eq:msvd-encoder}
\boldsymbol{v}_i = \boldsymbol{v}_{vid}, \quad
\boldsymbol{s}_i = \concat{\left(\boldsymbol{s}_t, \boldsymbol{cls}_{img}\right)}. 
\end{equation}
MSR-VTT dataset is relatively complicated compared to MSVD dataset. Thus, extra features are utilized in the experiments on MSR-VTT dataset: 
\begin{equation}\label{eq:msrvtt-encoder}
\boldsymbol{v}_i = \concat{\left(\boldsymbol{v}_{vid}, \boldsymbol{v}_{img}\right)}, 
\boldsymbol{s}_i = \concat{\left(\boldsymbol{s}_t, \boldsymbol{cls}_{vid}, \boldsymbol{cls}_{img}\right)}. 
\end{equation}
\subsection{Semantic-GRU-based Decoder with Variational Dropout and Layer Normalization} \label{vnsgru}
Traditional RNNs often suffer from serious overfitting on training set which deteriorates their performance on inference. 
Variational dropout is proposed in \cite{DBLP:conf/nips/GalG16} based on the mathematical grounding in deep Gaussian process. 
It has been widely applied in RNNs to prevent overfitting but it slows down the training process. 
Layer normalization has been proved to be very effective in stabilizing internal state dynamics in recurrent networks and Transformer \cite{Ba2016LayerN}. 
In the consideration of preventing overfitting as well as time and energy efficiency, 
variational dropout and layer normalization are embedded in our decoder together. 

The traditional gated recurrent unit (GRU) which has one fewer gate than LSTM and is capable of learning to acquire temporal dependencies across various scales \cite{DBLP:conf/emnlp/ChoMGBBSB14,chung2014empirical}. 
The unit consists of two gates: an update gate $\boldsymbol{z}_t$ and a reset gate $\boldsymbol{r}_t$\eqref{eq:UpdateResetGate}.  
Suppose we have input $\boldsymbol{x}_t \in \mathbb{R}^{n_x}$ and previous hidden state $\boldsymbol{h}_{t-1} \in \mathbb{R}^{n_h}$ at time step $t-1$, 
and then $\boldsymbol{z}_t$ and $\boldsymbol{r}_t$ can be computed as follows: 
\begin{equation}
\boldsymbol{z}_t = \sigma\left(\mathbf{W}_z \boldsymbol{x}_t + \mathbf{U}_z \boldsymbol{h}_{t-1}\right),\quad
\boldsymbol{r}_t = \sigma\left(\mathbf{W}_r \boldsymbol{x}_t + \mathbf{U}_r \boldsymbol{h}_{t-1}\right) ,
\label{eq:UpdateResetGate}
\end{equation}
where 
$\mathbf{W}_z \in \mathbb{R}^{n_h \times n_x}$, 
$\mathbf{U}_z \in \mathbb{R}^{n_h \times n_h}$, 
$\mathbf{W}_r \in \mathbb{R}^{n_h \times n_x}$ and
$\mathbf{U}_r \in \mathbb{R}^{n_h \times n_h}$. 
The candidate activation $\tilde{\boldsymbol{h}}_t$ 
is computed as follows:
\begin{equation}
\tilde{\boldsymbol{h}}_t =
\tanh\left(\mathbf{W}\boldsymbol{x}_t+\boldsymbol{r}_t\odot\left(\mathbf{U}\boldsymbol{h}_{t-1}\right)\right),
\label{candidateactivation}
\end{equation}
 where $\mathbf{W} \in \mathbb{R}^{n_h\times n_x}$, $\mathbf{U} \in \mathbb{R}^{n_h\times n_h}$ and $\odot$ denotes element-wise multiplication. 
 Unlike the popular version of the candidate activation
$\tilde{\boldsymbol{h}}_t = \tanh{\left(\mathbf{W}\boldsymbol{x}_t+\mathbf{U}\left(\boldsymbol{r}_t\odot \boldsymbol{h}_{t-1}\right)\right)}$
\cite{chung2014empirical}, the original approach \cite{cho2014on} is chosen to compute $\tilde{\boldsymbol{h}}_t$ for the sake of consistency. 
The activation of the gated recurrent unit at time $t$ is a linear interpolation between the previous activation 
$\boldsymbol{h}_{t-1}$ and the candidate activation $\tilde{\boldsymbol{h}}_{t}$:
\begin{equation}
\boldsymbol{h}_t = \left(1-\boldsymbol{z}_t\right)\boldsymbol{h}_{t-1}+\boldsymbol{z}_t\tilde{\boldsymbol{h}}_t\label{activation}.
\end{equation}
Inspired by \cite{DBLP:conf/cvpr/GanGHPTGCD17}, the weights $\mathbf{W}, \mathbf{U}, \mathbf{V}$ are transformed into 
semantics-dependent weight matrices to improve the quality of the generated captions as following shows. 
 \begin{IEEEeqnarray}{ll}
\hat{\boldsymbol{x}}_{t,\star} =& 
\mathbf{W}_{\star1}\boldsymbol{s} \odot \mathbf{W}_{\star2}\boldsymbol{x}_t, 
\label{candidatex} \\
\hat{\boldsymbol{h}}_{t,\star} =& 
\mathbf{U}_{\star1}\boldsymbol{s} \odot \mathbf{U}_{\star2}\boldsymbol{h}_t, 
\label{candidateh} \\
\hat{\boldsymbol{v}}_{\star} =&
\mathbf{V}_{\star1}\boldsymbol{s} \odot \mathbf{V}_{\star2}\boldsymbol{v},
\label{candidatev} \\
\boldsymbol{z}_t =& 
\sigma{\left(\mathbf{W}_{z3}\hat{\boldsymbol{x}}_{t,z}+\mathbf{U}_{z3}\hat{\boldsymbol{h}}_{t-1,z}+\mathbf{V}_{z3}\hat{\boldsymbol{v}}_{z}\right)}, 
\label{updategate2}\\
\boldsymbol{r}_t =&
\sigma{\left(\mathbf{W}_{r3}\hat{\boldsymbol{x}}_{t,r}+\mathbf{U}_{r3}\hat{\boldsymbol{h}}_{t-1,r}+\mathbf{V}_{r3}\hat{\boldsymbol{v}}_{r}\right)},
\label{resetgate2}\\
\tilde{\boldsymbol{h}}_t =&
\tanh{\left(\mathbf{W}_{h3}\hat{\boldsymbol{x}}_{t,h}+\boldsymbol{r}_t\odot\left(\mathbf{U}_{h3}\hat{\boldsymbol{h}}_{t-1,h}\right)+\mathbf{V}_{h3}\hat{\boldsymbol{v}}_{h}\right)},
\label{candidateactivation2} \\
\boldsymbol{h}_t =&
\left(1-\boldsymbol{z}_t\right)\boldsymbol{h}_{t-1}+\boldsymbol{z}_t\tilde{\boldsymbol{h}}_t
\label{activation2},
 \end{IEEEeqnarray}
 where $\Psi_{\star 1}\in \mathbb{R}^{n_f \times n_s}$, $\mathbf{W}_{\star2}\in \mathbb{R}^{n_f \times n_x}$, 
 $\mathbf{U}_{\star2} \in \mathbb{R}^{n_f \times n_h}$, $\mathbf{V}_{\star2} \in \mathbb{R}^{n_f \times n_v}$, 
 $\Psi_{\star3} \in \mathbb{R}^{n_h\times n_f}$, 
 $\star \in \{z, r, h \}$, $\Psi \in \{\mathbf{W}, \mathbf{U}, \mathbf{V}\}$.
 
With the consideration of preventing overfitting and training efficiency, layer normalization and variational dropout are applied to the semantic GRU.
 \begin{IEEEeqnarray}{ll}
\hat{\boldsymbol{x}}_{t,\star} &= \mathbf{W}_{\star1}\left(\boldsymbol{s}\odot\boldsymbol{m}_{\star,s}\right) \odot \mathbf{W}_{\star2}\left(\boldsymbol{x}_t\odot\boldsymbol{m}_{\star,x}\right),         \label{candidatex3} \\
\hat{\boldsymbol{h}}_{t,\star} &= \mathbf{U}_{\star1}\left(\boldsymbol{s}\odot\boldsymbol{m}_{\star,s}\right) \odot \mathbf{U}_{\star2}\left(\boldsymbol{h}_t\odot\boldsymbol{m}_{\star,h}\right),      \label{candidateh3} \\
\hat{\boldsymbol{v}}_{\star} &=\mathbf{V}_{\star1}\left(\boldsymbol{s}\odot\boldsymbol{m}_{\star,s}\right) \odot \mathbf{V}_{\star2}\left(\boldsymbol{v}\odot\boldsymbol{m}_{\star, v}\right),        \label{candidatev3} \\
\boldsymbol{z}_t &= \sigma{\left(\LN\left(\mathbf{W}_{z3}\hat{\boldsymbol{x}}_{t,z}+\mathbf{U}_{z3}\hat{\boldsymbol{h}}_{t-1,z}+\mathbf{V}_{z3}\hat{\boldsymbol{v}}_{z}\right)\right)},   \label{updategate3}\\
\boldsymbol{r}_t &=\sigma{\left(\LN\left(\mathbf{W}_{r3}\hat{\boldsymbol{x}}_{t,r}+\mathbf{U}_{r3}\hat{\boldsymbol{h}}_{t-1,r}+\mathbf{V}_{r3}\hat{\boldsymbol{v}}_{r}\right)\right)},  \label{resetgate3}\\
\tilde{\boldsymbol{h}}_t &=\tanh{\left(\LN\left(\mathbf{W}_{h3}\hat{\boldsymbol{x}}_{t,h}+\boldsymbol{r}_t\odot(\mathbf{U}_{h3}\hat{\boldsymbol{h}}_{t-1,h})+\mathbf{V}_{h3}\hat{\boldsymbol{v}}_{h}\right)\right)}, \nonumber\\
&\label{candidateactivation3} \\
\boldsymbol{h}_t &=\left(1-\boldsymbol{z}_t\right)\boldsymbol{h}_{t-1}+\boldsymbol{z}_t\tilde{\boldsymbol{h}}_t      \label{activation3},
 \end{IEEEeqnarray}
 where $\boldsymbol{m}$ is time-invariant dropout mask and $\LN$ denotes layer normalization. 
Inspired by the ``Go deeper, not wider'' principle of philosophy\cite{DBLP:conf/cvpr/SzegedyLJSRAEVR15,DBLP:conf/cvpr/HeZRS16}, we stack two variational-normalized semantic GRU (VNS-GRU) layers together and reduce the internal embedding dimension $n_f$.  
As a result of it, the model has stronger decoding ability than the one-layer model, even though it has fewer parameters than the latter one. 

\subsection{Selection of Model for Testing: Comprehensive Selection Method}\label{section:sel}
Given $n$ metrics $M=[m_0, m_1,..., m_{n-1}]$ with the corresponding weights $W=[w_0,w_1,...,w_{n-1}]$, 
the overall performance $o$ of a model can be evaluated as follows: 
\begin{equation}
o = \sum_{i} \frac{w_i \cdot v_i}{b_i}, \label{eq:sum}
\end{equation}
where $v_i$ is the value of metric $m_i$ and $b_i$ is by far the best value for metric $m_i$. 
At the end of each epoch, $v_i$ is computed based on the output of the model on the validation set. 
$b_i$ for each metric and $o_b$ are updated subsequently if necessary. 
A checkpoint of the model is saved whenever $o_b$ is updated during training process. 

In our method, the performance of a model is evaluated by $n$ metrics on the validation set with predefined weights $W$. 
If $M=[Cross Entropy]$, then the best model is chosen solely based on cross entropy loss for testing. 
Our method can be embeded into an end-to-end training framework to assess a checkpoint by arbitrary number of metrics. 
Most of the existing selection methods can be regarded as special cases of this method. 

\begin{figure}[thb]
\begin{center}
\includegraphics[scale=0.26]{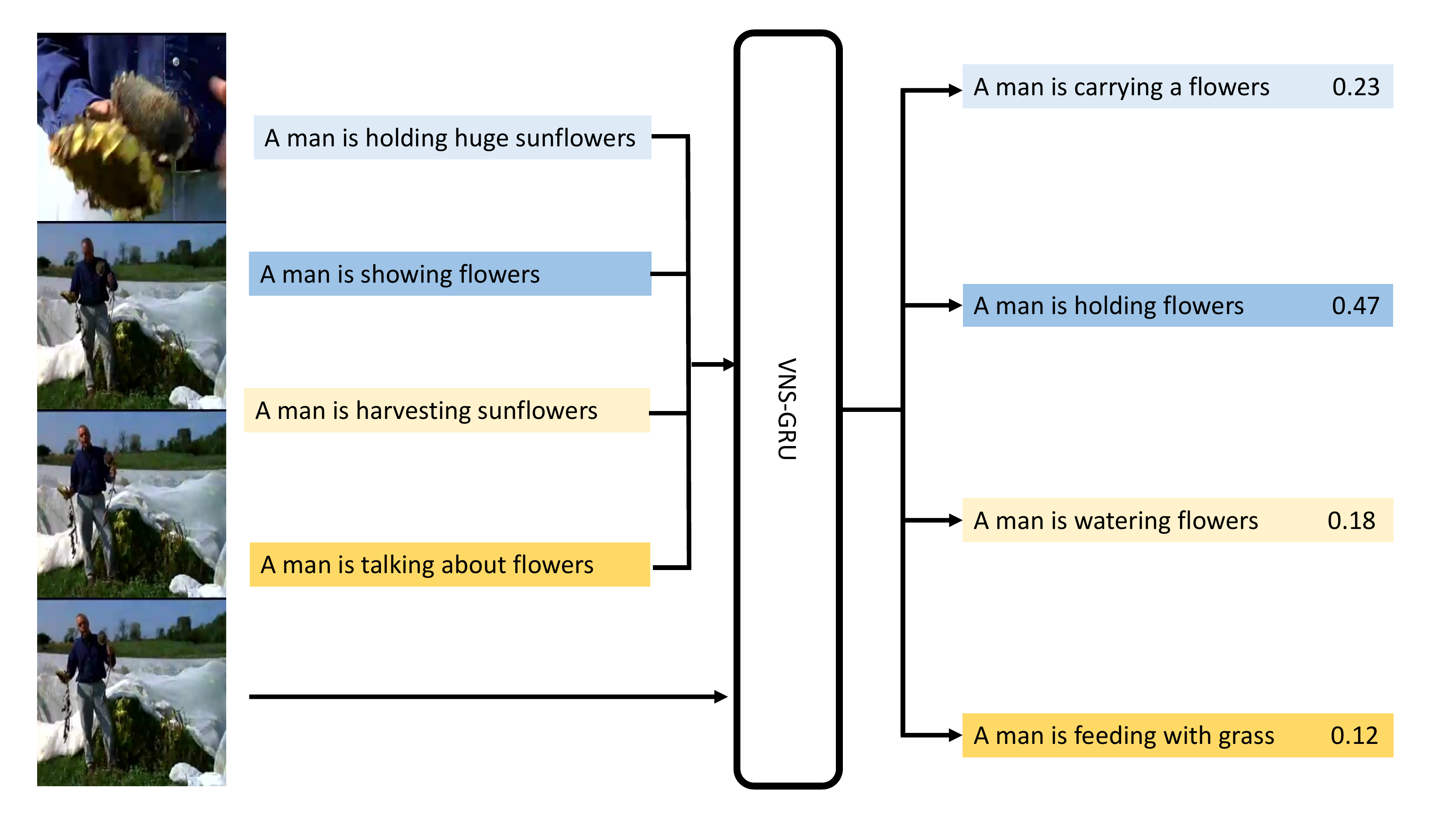}
{\caption{VNS-GRU stands for semantic gated recurrent unit model with variational dropout and layer normalization. 
Video features are fed into our model and the model is optimized by different annotations with corresponding weights 
and it is called \textit{professional learning}.}\label{vnsgrupl}}
\end{center}
\end{figure}

\subsection{Training Strategy: \textit{Professional Learning}}
For previous training strategies, models are optimized on all training samples equally which leads to ``absolute equalitarianism'' implicitly. 
``Absolute equalitarianism'' in video captioning task is a phenomenon that a model's knowledge for the common part of all training samples is enhanced iteratively 
and the model is inclined to forget advanced words and complicated language grammar since they vary too much from sentence to sentence. 
The intersection of human annotations for a video usually consists of limited number of frequent vocabulary and elementary language grammar. 
It partially explains why an ordinary video captioning model cannot generate a sentence that is competitive with human annotations. 

Inspired by the higher education system in the real word, we propose a novel training method called \textit{professional learning} (Alg. \ref{alg:prof}). 
University students get a liberal education to consolidate and widen their basic knowledge and skills first and then choose a particular specialty to develop their professional skills. 
Similarly, in \textit{professional learning}, a model will be trained by optimizing losses computed with training samples equally in the first phase, 
which is called \textit{teacher forcing} or \textit{general learning}.
In the second phase, $n$ annotations $\mathbf{A}^{(k)} = [\boldsymbol{a}^{(k)}_0, \boldsymbol{a}^{(k)}_1,...,\boldsymbol{a}^{(k)}_{n-1}]$ are sampled for the video $k$. 
The possibility distributions 
$\mathbf{P}^{(k)} = [\boldsymbol{p}^{(k)}_{0}, \boldsymbol{p}^{(k)}_{1}, ..., \boldsymbol{p}^{(k)}_{n-1}]$ 
for each token in the generated captions 
$\mathbf{C}^{(k)} = [\boldsymbol{c}^{(k)}_0, \boldsymbol{c}^{(k)}_1,...,\boldsymbol{c}^{(k)}_{n-1}]$ are computed by the model, 
where $\boldsymbol{c}^{(k)}_{i}$ is guided by $\boldsymbol{a}^{(k)}_i$. 
Cross entropy loss $\boldsymbol{l}^{(k)}$ is produced for pairs of possibility distribution $\boldsymbol{p}^{(k)}_{i}$ and human annotation $\boldsymbol{a}^{(k)}_{i}$: 
\begin{equation}
\boldsymbol{l}^{(k)}=[l^{(k)}_0, l^{(k)}_1, ..., l^{(k)}_{n-1}]^\intercal, 
l^{(k)}_i = \mean{\left(-\boldsymbol{a}^{(k)}_{i}\log{\boldsymbol{p}^{(k)}_{i}}\right)}. 
\label{eq:xesample}
\end{equation}

The cross entropy loss $\boldsymbol{l}^{(k)}$ with weights $\boldsymbol{\beta}^{(k)}$ is utilized to optimize a video captioning system 
and the weighted loss is formulated as: 
\begin{equation}
Loss\left(\mathbf{A}, \mathbf{S}, \mathbf{V}; \Theta\right)= \frac{1}{bs}\sum_{i=0}^{bs-1}\boldsymbol{\beta}^{(i)} \boldsymbol{l}^{(i)}, \label{eq:loss}
\end{equation}
where $\mathbf{A}$, $\mathbf{S}$, ${\mathbf{V}}$ and $\Theta$ denote human annotations, semantic information, visual features, and model parameters respectively. 

Given loss $l^{(k)}_i$, the corresponding annotation $\boldsymbol{a}^{(k)}_i$ and the corresponding caption $\boldsymbol{c}^{(k)}_i$, 
small loss $l^{(k)}_i$ indicates the model ''is good at'' generating a caption $\boldsymbol{c}^{(k)}_i$ which is similar to $\boldsymbol{a}^{(k)}_i$; 
large loss $l^{(k)}_i$ indicates the model is inclined to generate a caption which is dissimilar to $\boldsymbol{a}^{(k)}_i$. 
The former property is called a strength of the model and the latter is called a weakness. 
The generated caption $\boldsymbol{c}^{(k)}_i$ has large weight $\beta^{(k)}_i$ in optimization 
if the corresponding loss $l^{(k)}_i$ is small; 
$\boldsymbol{c}^{(k)}_i$ has small $\beta^{(k)}_i$ if $l^{(k)}_i$ is large. 
Strengths of a model are enhanced and weaknesses are bypassed in this way. 
The model is able to learn unique words and advanced grammar rules for they pay more attention on the samples they do well in. 
Larger size of vocabulary and more diverse sentence structures are expected to be observed in the captions produced by the model trained in this way. 
Weights $\boldsymbol{\beta}$ consists of two parts: 
\begin{equation}
\begin{aligned}
\boldsymbol{\beta}^{(i)}&=
\gamma\softmax{\left(-\boldsymbol{l}^{(i)}\right)}\\
&+\left(1-\gamma\right)\softmax{\left(-\abs\left(\boldsymbol{len}^{(i)}-\bar{len}\right)\right)}, \label{eq:lossBeta}
\end{aligned}
\end{equation}
where $\gamma$ is a hyper-parameter to modulate the balance between the cross-entropy-related and length-related probability distribution. 
The first part is probability distribution determined by $\softmax$ value of cross entropy so that samples with higher loss values have lower probability and vice versa.  
The second part is $L1$ distance between machine-generated and average sentence length in a dataset which is to encourage the generation of average-length sentences.

It is easier to generate short sentence with small loss than to generate long sentence with the same loss because of accumulation of errors in RNN. 
Without the second term in \eqref{eq:lossBeta}, short sentences may gain unproportionally large weights $\boldsymbol{\beta}$ in optimization.
If $\gamma$ is close to 1, a model is likely to generate short and simple captions for it is relatively easy to fit on such kind of captions and loss values for those samples are small; vice versa.

\begin{algorithm}[htb]
\caption{Professional Learning Algorithm}
\label{alg:prof}
\KwData{inputs $\mathbf{i}$, human annotations $\mathbf{a}$, total training epoch $epoch_{total}$ and the switch point $epoch_{sw}$}
$idx_{cur} \gets 0$\;
\While{$idx_{cur} < epoch_{total}$}{
\eIf(\tcp*[h]{\small\textit{general learning} scheme}){$idx_{cur} < epoch_{sw}$}{
Optimize the model by each video and annotation pair equally\;
}(\tcp*[h]{\small\textit{professional learning} scheme}){
Compute and optimize the model by weighted loss function with $idx_{cur}$,  inputs $\mathbf{i}$ and human annotation $\mathbf{a}$\;

}
$idx_{cur} = idx_{cur} + 1$\;
}
\end{algorithm}

For the sake of a smooth transition, a proper schedule for the sampling number $n$ is needed. Examples of the schedule for $n$ can be as follows:
\begin{enumerate}
\item Fixed sampling size: $n=c\in \mathbb{Z}^+$. 
\item Exponential schedule: 
\[n=\left\{
\begin{array}{ll}
1, &\text{if } idx_{cur} <= epoch_{sw}, \\
b^{\floor\left(\left(idx_{cur}-epoch_{sw}\right)/\sigma\right)}, &\text{if } idx_{cur} > epoch_{sw}.
\end{array} \right.,\] where $\sigma \in \mathbb{Z}^+$ is a hyper-parameter that depends on the estimated rate of convergence. 
\end{enumerate}

\begin{figure}[htb]
\begin{center}
\begin{tabular}{l}\toprule
\includegraphics[width=240pt]{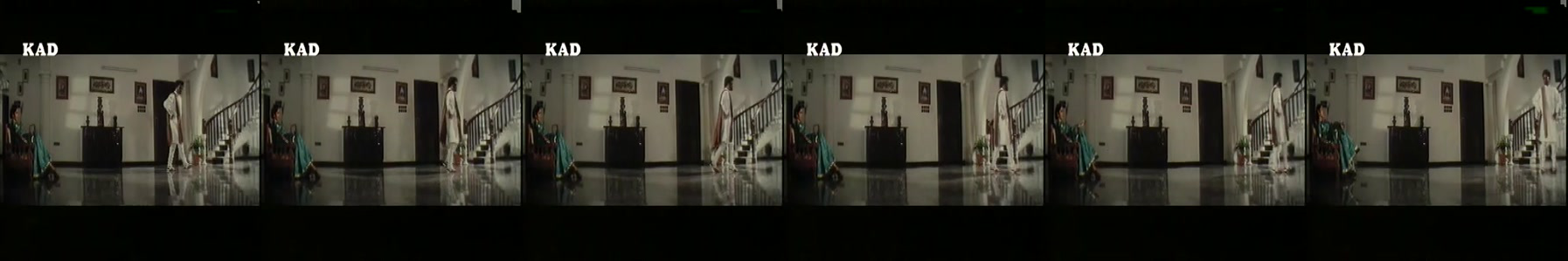} 	\\
GT: (1) a man is walking  (2) the person is walking			\\	\midrule
\includegraphics[width=240pt]{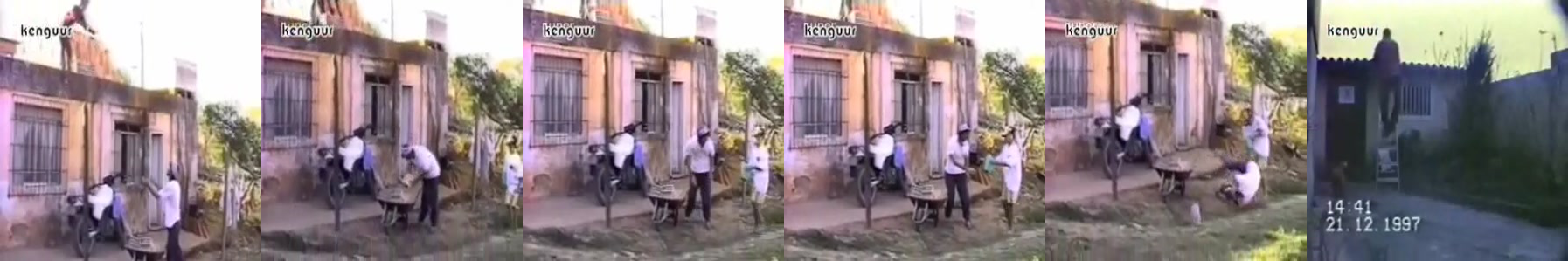} 	\\
GT: (1) a man falls down  (2) a man falls over		\\	\midrule
\includegraphics[width=240pt]{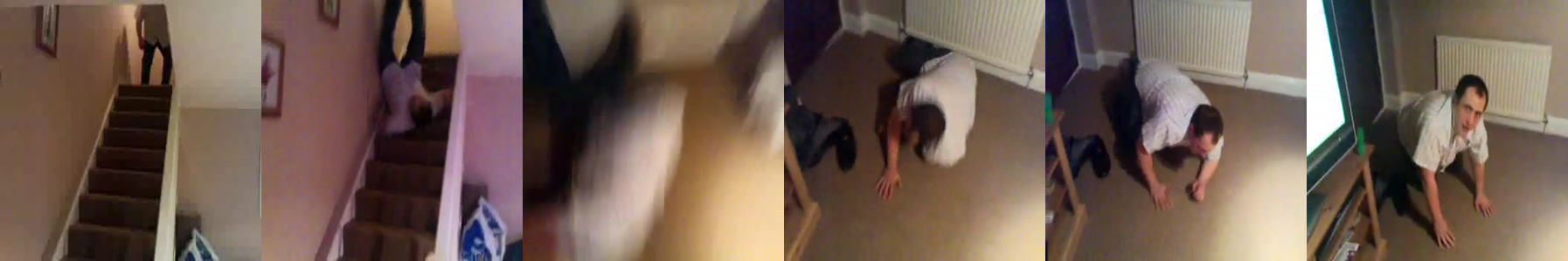}\\ 
GT: (1) a man falling down stairs (2) a man who is drunk falls down	\\ \bottomrule 
\end{tabular}
{\caption{Duplicate captions for different videos. 
GT is short for ground truth. 
The machine-generated captions for these three video clips are identical: ``a man is running''. 
Two actions ``walk'' and ``fall down'' are recognized as ``run'' incorrectly. }\label{duplicatedCaption}}
\end{center}
\end{figure}

\section{Experiment} 
We implement our models and perform experiments under the TensorFlow framework. 
Its source code can be found on GitHub
\footnote{\tiny\url{https://github.com/WingsBrokenAngel/delving-deeper-into-the-decoder-for-video-captioning}}.
\subsection{Dataset}
Two popular datasets for video captioning in recent research are used in experiments, 
MSVD and MSR-VTT.
\subsubsection{MSVD Dataset}
The MSVD dataset consists of 1970 short video snippets.
Each video clip has length of 10s on average and each of them corresponds to 40 English human annotations in MSVD \cite{DBLP:conf/iccv/GuadarramaKMVMDS13}. 
The average number of words for each annotation in the dataset is around 7.1 which implies those annotations are composed of relatively short and simple sentences.  
Following the split setting in \cite{DBLP:conf/iccv/GuadarramaKMVMDS13, DBLP:conf/iccv/HoriHLZHHMS17, DBLP:conf/naacl/WangWW18, 
DBLP:journals/ijon/WuWCSSW18, DBLP:conf/cvpr/AafaqALGM19, DBLP:conf/cvpr/PeiZWKST19}, 
we take 1200 video clips as training set, 100 video snippets as validation set and 670 clips as testing set. 
We tokenize 80839 English sentences and obtain vocabulary with 12596 English words from the training set. 
\subsubsection{MSR-VTT Dataset}
The MSR-VTT 1.0 dataset is composed of 10000 short video clips, which are divided into twenty predefined categories \cite{DBLP:conf/cvpr/XuMYR16}. 
Each video snippet has roughly 20 English descriptions. 
The average number of words for each annotation in the dataset is around 9.3 which implies that human annotations in MSR-VTT are more complicated than those in MSVD. 
We follow data split setting provided by MSRA and take 6513 video clips as training set, 497 video clips as validation set and 2990 video clips as testing set.
We tokenize 200000 English human annotations and obtain vocabulary with 13796 English words only from the training set.

\subsection{Model Architecture}
In the encoder, ResNeXt-101 with 64 paths in each block pre-trained on ImageNet is used as our frame-level feature generator \cite{DBLP:conf/cvpr/XieGDTH17}. 
The 2048-dim feature map for each frame is taken from the output of the global pooling layer in ResNeXt. 
We also collect the probability distribution of classification for each frame and apply average pooling operation on the frames from the same video. 
The averaged probability distribution, a 1000-dim vector for each video, is used as part of the semantic features.  
We choose Efficient Convolutional Network (ECN) \cite{DBLP:conf/eccv/ZolfaghariSB18} pre-trained on Kinetics-400 as our video-level feature generator. 
The feature map for each video clip is taken from the output of concatenation of the global pooling layers in ECN. 
Both feature maps from ResNeXt and ECN are scaled to $[0,1]$. 
The probability distribution for actions from ECN, which is a 400-dim vector for each video clip, is taken as (a part of) semantic information.
In addition, we select 300 key words from dataset vocabulary as tags (semantic clues) for each video. 
Our tagging network is trained on training and validation sets and is utilized to predict tags for each video and each of them is a 300-dim vector. 
In all, we have video feature with dimension 1536 and semantic feature with dimension 1300 for each video segment in MSVD dataset \eqref{eq:msvd-encoder}, 
and video feature with dimension 3584 and semantic feature with dimension 1700 for each of those in MSR-VTT dataset \eqref{eq:msrvtt-encoder}. 
Note that we reuses the video and semantic features in SAM-SS\cite{2019arXiv190900121C} for the sake of convenience. 

For MSVD dataset, our model has settings as follows: $n_v=1536$, $n_s=1300$, $n_f=64$, $n_h=n_x=512$ and $\gamma=0.8$.
For MSR-VTT dataset, our model has settings as follows: $n_v=3584$, $n_s=1700$, $n_f=128$, $n_h=n_x=1024$ and $\gamma=0.4$. 
Given that all the other hyper-parameters are equal, 
the number of parameters of a model with $n_f=\frac{n_h}{8}$ is about one-eighth of the one with $n_f=n_h$, 
while the performance of the former can be comparable with the latter.

\subsection{Training Detail} \label{section:train}
The model is trained for 50 epochs on MSVD dataset and for 80 epochs on MSR-VTT dataset. And the best model is chosen based on the performance on the validation set as described in Section \ref{section:sel}. 
In our experiment, B4, CIDEr, METEOR and ROUGE\-L are used to evaluate the performance of a model and the weights for each of them are set to 0.25\eqref{eq:overall}. 
\begin{equation}
overall_i=\left(\frac{B4_i}{B4_b}+\frac{C_i}{C_b}+\frac{M_i}{M_b}+\frac{R_i}{R_b}\right)/4,\label{eq:overall}
\end{equation}
where $m_b$ is the current best score on metric $m \in [B4, C, M, R]$\cite{2019arXiv190900121C} and subscript $i$ denotes checkpoint $i$. 
The training strategy is switched from \textit{general learning} scheme to \textit{professional learning} scheme at 16th epoch. 
For MSVD, the sampling number $n$ of annotations for each video is fixed to 16. 
The sampling number $n$ in MSR-VTT is computed as follows: 
\begin{equation}
n=2^{\floor\left(e/16\right)},\label{samplingNumber}
\end{equation}
where $e$ is the epoch index during training. 
The reason for the choice of the sampling schedules is described in Section \ref{section:sampling}. 
A GeForce GTX 1080 Ti is utilized to speed up the training process for each of those experiments. 
The model finishes its training on MSVD dataset within two hours and on MSR-VTT dataset within six hours. 
Our models are optimized by Adam Algorithm with initial learning rate $2\times 10^{-4}$ and global norm gradient clip of 40. 
We use a weight decay of 0.861 every 1000 steps for MSVD and 0.9455 every 1000 steps for MSR-VTT. 
\subsection{Ablation Study}
\subsubsection{Influence of Sampling Schedule in \textit{Professional Learning}} \label{section:sampling}
We perform experiments with different fixed sampling size $n\in\{2, 4, 8, 16\}$ and exponential sampling schedule on MSVD and MSR-VTT dataset. 
As demonstrated by Table \ref{table:samplesize}, on MSVD dataset, 
the best performance evaluated by overall score is obtained with $n=16$. 
It also can be inferred from Table \ref{table:samplesize} that, on MSR-VTT dataset, 
the best performance among different fixed sampling size is obtained with $n=8$ which is smaller than the one on MSVD dataset. 
The best performance on MSR-VTT among all schedules is obtained with exponential sampling schedule. 

The human annotations in MSVD dataset are simpler than those in MSR-VTT and 
the diversity of sentences in MSVD is less than that in MSR-VTT. 
For a video, annotations have more words and sentence structures in common in MSVD than in MSR-VTT. 
A model is able to achieve better performance with more attention focused on few sentences in MSVD 
but it needs to allocate its attention more evenly in MSR-VTT to have better metric values. 
Exponential schedule optimizes the model on different sampling size so that it helps the model learn better hidden patterns from diversified annotations in MSR-VTT. 
\begin{table}
\begin{center}
{\caption{Model performance with different fixed sampling number $n$ on MSVD and MSR-VTT dataset. 
EXP denotes exponential schedule\eqref{samplingNumber}. Size denotes sampling size.}
\label{table:samplesize}}
\begin{tabular}{ccccccc}
\toprule
DS		&Size &   B4 	&    C 			&  M  			&  R   			&   Overall\eqref{eq:overall}\\
\midrule
\multirow{5}{*}{MSVD}
		&2      &64.0    		&117.8   		&41.4    		&79.3			&0.978			\\
		&4		&62.9			&118.9			&41.4			&79.2			&0.975			\\
		&8		&64.5			&121.0			&41.6			&79.1			&0.987			\\
		&16		&\textbf{66.5}	&\textbf{121.5}	&\textbf{42.1}	&\textbf{79.7} 	&\textbf{1.000}	\\ 
		&EXP	&64.1			&119.1			&41.5			&79.1 			&0.981			\\ 
		\midrule
\multirow{5}{*}{MSR-VTT}
		&2        	&44.5	    	&51.8	  		&29.4    		&62.9			& 0.980			\\
		&4			&45.0			&51.7			&29.4			&62.9			& 0.982			\\
		&8			&\textbf{46.1}	&51.4			&29.3			&62.8			& 0.985			\\
		&16			&45.8			&49.6			&28.6			&62.5		 	& 0.968			\\
		&EXP		&45.3			&\textbf{53.0}	&\textbf{29.9}	&\textbf{63.4}	&\textbf{0.996}	\\
\bottomrule
\end{tabular}
\end{center}
\end{table}

\subsubsection{Effectiveness of Components}
As shown by Table \ref{table:performancecomponents}, we have high-level baseline model, by virtue of enhanced semantic features and highly qualified visual features. 
If cross entropy is used to select the best model for testing, a low-quality model will be chosen. 
The chosen model will have better performance if BLEU-4 is used for selecting. 
Comprehensive Selection Method is able to find the model with the satisfying overall performance based on the metric values in validation set.
The combination of variational dropout and layer normalization makes a profound impact on the model performance in MSVD and MSR-VTT. 
Our model is improved by \textit{professional learning} significantly on both datasets which proves the validity of the proposed method, 
as demonstrated by Table \ref{table:performancecomponents}. 
The sampling schedules for \textit{professional learning} are described in Section \ref{section:train}.
In overall, the performance of our model is improved with the increase in the number of modules or methods.

\begin{table}
\begin{center}
{\caption{The results of the experiments performed on MSVD and MSR-VTT with and without some components. 
SEL, VD, LN and PL are the abbreviations for 
selection method described in Section \ref{section:sel}, variational dropout and \textit{professional learning}, respectively. 
XE in the column of SEL denotes cross entropy loss which is used for selecting the model. 
BLEU4 means that BLEU-4 is used to select the model for testing.
}\label{table:performancecomponents}}
\begin{tabular}{cccccccc}
\toprule
Dataset 						&SEL		&VD\&LN			&PL			&B4				&C					&M				&R 				\\
\midrule
\multirow{5}{*}{\text{\tiny{MSVD}}}	
						&XE 		&$\times$	&$\times$		& 60.9 			& 100.3 			& 38.7 			& 76.7 			\\
						&BLEU4		&$\times$	&$\times$		& 57.4			& 104.4				& 38.3			& 75.7			\\
						&$\surd$	&$\times$	&$\times$		& 57.4			& 104.4				& 38.3			& 75.7 			\\
						&$\surd$	&$\surd$	&$\times$		& 64.3			& 117.5				& 40.9			& 78.9 			\\
						&$\surd$	&$\surd$	&$\surd$		&\textbf{66.5}	&\textbf{121.5}		&\textbf{42.1}	&\textbf{79.7}	\\ \midrule
\multirow{5}{*}{\text{\tiny{MSR-VTT}}}
						&XE 		&$\times$	&$\times$		& 40.0 			& 44.3 				& 27.0 			& 60.4 			\\
						&BLEU4		&$\times$	&$\times$		& 40.0			& 46.6				& 27.5			& 60.4			\\
						&$\surd$	&$\times$	&$\times$		& 41.5			& 48.8				& 28.3			& 61.5			\\
						&$\surd$	&$\surd$	&$\times$		& 44.0			& 50.1				& 28.7			& 62.3			\\
						&$\surd$	&$\surd$	&$\surd$		&\textbf{45.3}	&\textbf{53.0}		&\textbf{29.9}	&\textbf{63.4}	\\
\bottomrule
\end{tabular}
\end{center}
\end{table}

\subsubsection{Diversity of the Generated Captions}
Poor vocabulary and repeated captions are two serious problems in video captioning task. 
We count the number of distinct sentences and the number of distinct words in the captions generated by our model for MSVD and MSR-VTT test sets respectively. 

As demonstrated by Table \ref{table:unique}, for the VS-GRU baseline, 
only $196/12596=1.6\%$ and $342/13796=2.5\%$ of all the vocabulary appears in the generated captions for the MSVD and MSR-VTT test sets respectively. 
For the same model, $670 / 326=2.1$ and $2990/793=3.8$ clips of video share the same caption on average in MSVD and MSR-VTT testing respectively. 
As shown in Fig. \ref{duplicatedCaption}, three video clips are described with identical captions: ``a man is running''. 
Layer normalization enlarges the vocabulary of the model by $13.3\%$ on MSVD test set and $13.5\%$ on MSR-VTT test set. 
\textit{Professional learning} enlarges vocabulary by $14.9\%$ on MSVD test set and $7.2\%$ on MSR-VTT test set. 
These two methods also increase the number of distinct sentences accordingly. 
Layer normalization stabilizes the internal states of the model so that the model is able to learn delicate pattern hidden in features. 
In this way, the model can generate different words in response to the slight changes brought by inputs. 
In some degree, \textit{professional learning} and layer normalization alleviate the problems of poor vocabulary and repeated captions in testing. 

\begin{figure}[htb]
\begin{center}
\begin{tabular}{l}\toprule
\includegraphics[width=240pt]{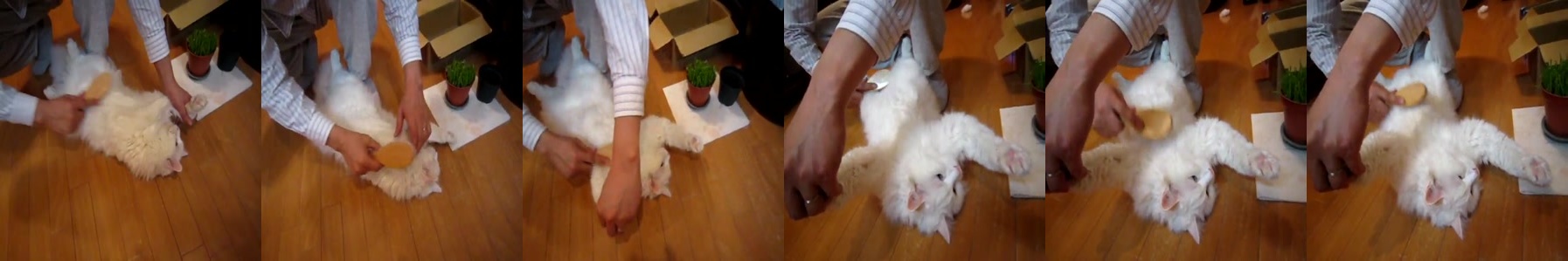} 	\\
without: a cat is {\color{red}playing}. with: a person is {\color{green}cleaning} a cat.	\\ 
GT: a man is brushing a cat. 							\\	\midrule
\includegraphics[width=240pt]{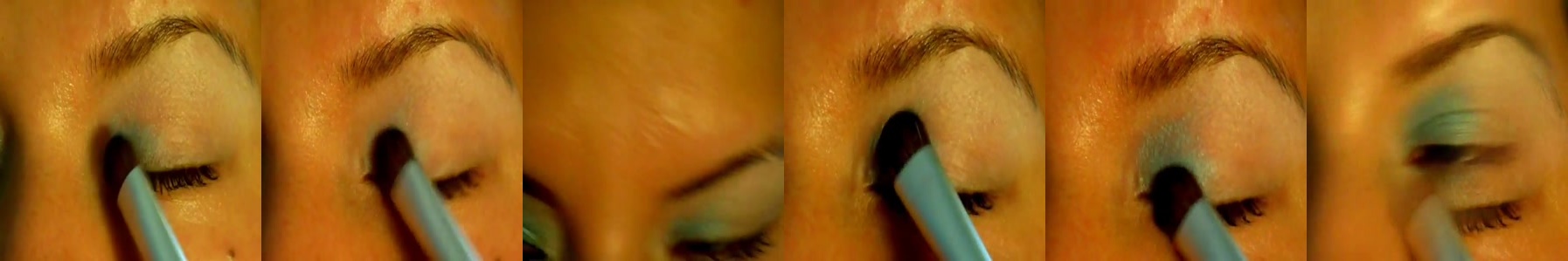}	\\ 
without: a woman is applying {\color{red}a eye}. 		\\
with: a woman is {\color{green}applying makeup}. 		\\
GT: a woman is putting makeup. 				\\ 	\midrule
\includegraphics[width=240pt]{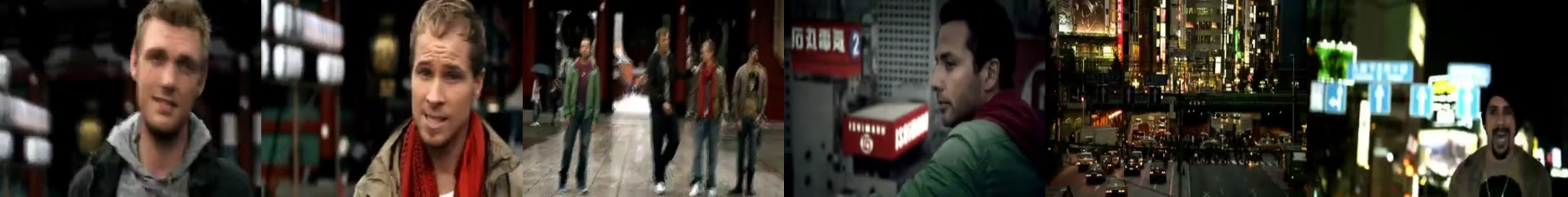}	\\ 
without: {\color{red}a man} is singing and dancing.		\\
with: {\color{green}a group of} people are singing and dancing. \\ 
GT: {\color{red}group of} people {\color{red}of} singing and walking on streets. \\ \midrule
\includegraphics[width=240pt]{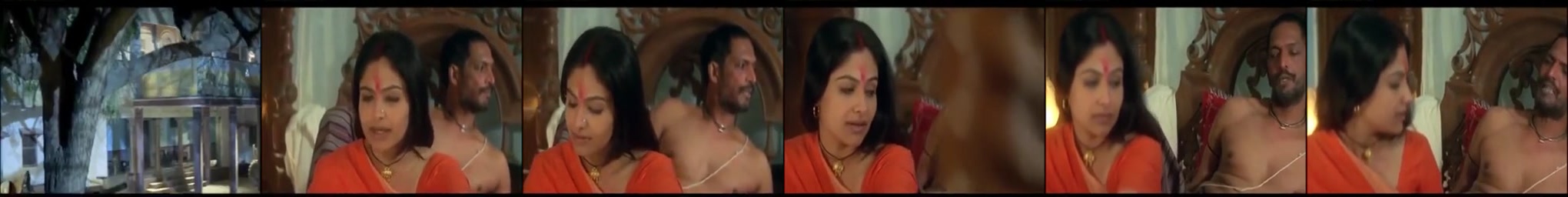}\\ 
without: a {\color{red}woman} and a woman are talking. \\
with: a {\color{red}woman} and a woman are talking to {\color{green}each other}. \\
GT: a man and woman sitting in bed talking in a {\color{red}fireign} language.	\\ \bottomrule
\end{tabular}
{\caption{Comparison between models with or without \textit{professional learning}. 
GT is short for ground truth. Errors or mistakes in captions are indicated by red color, such as ``playing''. 
The model trained by \textit{professional learning} learns to describe video clips more accurately. 
It is able to make use of advanced words or phrases, indicated by green color, to generate sentences.}\label{imageCaptions}}
\end{center}
\end{figure}

Sometimes, our method can generate a caption which is competitive with or even better than a human annotation. 
In the first video clip of Fig. \ref{imageCaptions}, our model uses ``cleaning'' to describe the action of human which expresses the purpose of ``brushing''. 
In the third video clip, ``a group of'', which is grammatically correct, is used by our model instead of ``group of'' in the ground truth, which is a syntax error. 

\begin{table}
\begin{center}
{\caption{Statistics data of distinct sentences and vocabulary size for models without certain component(s).
LN and PL denote layer normalization and \textit{professional learning} respectively. 
VS-GRU stand for a semantic GRU model with variational dropout.}\label{table:unique}}
\begin{tabular}{ccccc}
\toprule
Dataset						& LN			& PL 			& Distinct Sentences & Vocabulary Size\\
\midrule 		
\multirow{3}{*}{MSVD}	& $\times$		& $\times$		& 326						& 196				\\
						& $\surd $		& $\times$		& 376						& 222				\\
 						 & $\surd$		& $\surd$		&\textbf{397}				&\textbf{255}\\ \midrule
\multirow{3}{*}{MSR-VTT}	& $\times$		& $\times$		& 793						& 342 	\\ 
		 				& $\surd$		& $\times$		& 910						& 388				\\
 						& $\surd$		& $\surd$		&\textbf{915}				&\textbf{416}		\\
\bottomrule
\end{tabular}
\end{center}
\end{table}

\subsection{Comparison with Previous Models}
To demonstrate the superiority of our method, we list the performance of our model along with the previous state-of-the-art results from existing computer vision literature. 
Four metrics in natural language processing, called BLEU-4, CIDEr, METEOR and ROUGE-L, are applied to evaluate the performance of those models numerically. 
\subsubsection{Comparison on MSVD}
\begin{table}
\begin{center}
{\caption{The results of the models on MSVD. 
VNS-GRU is a model trained without $\boldsymbol{cls}_{img}$ 
while VNS-GRU$^r$ is a model trained with it \eqref{eq:msvd-encoder}. 
The rest configuration of two models is the same and is described in Section \ref{section:methods}. 
B4, C, M and R stand for BLEU-4, CIDEr, METEOR and ROUGE-L respectively.}
\label{table:performanceMSVD}}
\begin{tabular}{lccccr}
\toprule
Models                                          & B4    	& C  		& M 		& R 		& Overall\eqref{eq:overall}\\
\midrule
SCN\cite{DBLP:conf/cvpr/GanGHPTGCD17}        	& 51.1      & 77.7   	& 33.5     	& -   		&	-	\\
MTVC\cite{DBLP:conf/acl/PasunuruB17}            & 54.5      & 92.4   	& 36.0     	& 72.8    	& 0.837	\\
CIDEnt-RL\cite{DBLP:conf/emnlp/PasunuruB17}  	& 54.4     	& 88.6    	& 34.9      & 72.2   	& 0.821	\\
HATT\cite{DBLP:journals/ijon/WuWCSSW18}    		& 52.9     	& 73.8    	& 33.8      & -    		&	-	\\
ECN\cite{DBLP:conf/eccv/ZolfaghariSB18}         & 53.5    	& 85.8     	& 35.0     	& -     	& 	-	\\
GRU-EVE\cite{DBLP:conf/cvpr/AafaqALGM19}    	& 47.9     	& 78.1   	& 35.0      & 71.5    	& 0.773	\\
MARN\cite{DBLP:conf/cvpr/PeiZWKST19}          	& 48.6    	& 92.2   	& 35.1      & 71.9   	& 0.806	\\
SAM-SS\cite{2019arXiv190900121C}                & 61.8      & 103.0    	& 37.8      & 76.8   	& 0.910	\\
\midrule
VNS-GRU											& 66.3		& 116.0		& 41.4		& 79.1		& 0.982	\\
VNS-GRU$^r$                       				&\textbf{66.5}&\textbf{121.5}	&\textbf{42.1}&\textbf{79.7} &\textbf{1.000}\\
\bottomrule
\end{tabular}
\end{center}
\end{table}

To the best of our knowledge, our model outperforms all the previous models on all the metrics (Table \ref{table:performanceMSVD}).
SCN\cite{DBLP:conf/cvpr/GanGHPTGCD17} takes advantage of semantic information produced by video tagging network 
and ensemble the RNN weights with video tags. 
Its performance is evaluated on the ensemble of five models. 
Multi-task video captioning model (MTVC) \cite{DBLP:conf/acl/PasunuruB17} is trained with unsupervised video prediction task to learn resilient video encoder representation 
and language entailment task to produce logic-enhanced annotation decoder feature. 
CIDEnt-RL\cite{DBLP:conf/emnlp/PasunuruB17} is trained by reinforcement method using mixed-loss methods and entailment-enhanced reward 
which outperforms CIDEr-reward models. 
HATT \cite{DBLP:journals/ijon/WuWCSSW18} utilizes temporal features, motion features, audio information and semantic information by hierarchical attention-based fusion 
to generate captions for videos. 
Efficient Convolutional Network (ECN) \cite{DBLP:conf/eccv/ZolfaghariSB18} is an 3D convolutional network for video action classification task and meaningful video features outputted by it are fed into SCN to produce annotations for videos. 
In GRU-EVE \cite{DBLP:conf/cvpr/AafaqALGM19}, Hierarchical Short Fourier Transform is applied to frame-level features in order to derive high-quality temporal dynamics and 
rich high-level semantic information is obtained from an object detection model. 
MARN \cite{DBLP:conf/cvpr/PeiZWKST19} employs a memory block to store all the related visual and contextual information over the training set for each word.
SAM-SS \cite{2019arXiv190900121C} is trained by scheduled sampling method with the assistance of semantics.

The experiment results displayed in Table \ref{table:performanceMSVD} show that our model outperforms all the other methods on all the metrics with a large margin. 
Our model VNS-GRU$^r$ achieves gains over the previously best model SAM-SS by 7.6\% on BLEU-4, by 18.0\% on CIDEr, by 11.4\% on METEOR and by 3.8\% on ROUGE-L.

\begin{table}
\begin{center}
{\caption{The results of the models on MSR-VTT. 
$\star$ denotes that the model is from MSR-VTT Challenge 2017. 
VNS-GRU is trained by the features shown in \eqref{eq:msvd-encoder}. 
Subcript $r$ denotes feature $\boldsymbol{v}_{img}$ is used in training. 
Superscript $e$ and $r$ denote semantic feature $\boldsymbol{cls}_{vid}$ and $\boldsymbol{cls}_{img}$ are used in training respectively \eqref{eq:msrvtt-encoder}. 
}\label{table:performanceMSRVTT}
}
\begin{tabular}{lccccr}
\toprule
Models										&B4				&C 				&M 				&R 				& Overall\\
\midrule
v2t\_navigator$\star$           			& 40.8     		& 44.8    		& 28.2      	& 60.9   		& 0.911	\\
Aalto$\star$                        		& 39.8     		& 45.7    		& 26.9      	& 59.8   		& 0.894	\\
VideoLAB$\star$                  			& 39.1     		& 44.1    		& 27.7      	& 60.6   		& 0.893	\\
\midrule
CIDEnt-RL\cite{DBLP:conf/emnlp/PasunuruB17} & 40.5     		& 51.7    		& 28.4      	& 61.4   		& 0.945	\\
HACA\cite{DBLP:conf/naacl/WangWW18}         & 43.4     		& 49.7    		& 29.5			& 61.8  		& 0.963	\\
HATT\cite{DBLP:journals/ijon/WuWCSSW18}    	& 41.2     		& 44.7    		& 28.5      	& 60.7    		& 0.914	\\
GRU-EVE\cite{DBLP:conf/cvpr/AafaqALGM19}    & 38.3     		& 48.1    		& 28.4      	& 60.7    		& 0.914	\\
MARN\cite{DBLP:conf/cvpr/PeiZWKST19}        & 40.4    		& 47.1   		& 28.1      	& 60.7   		& 0.918	\\
TAMoE\cite{DBLP:conf/aaai/WangWZ0W19}      	& 42.2   		& 48.9   		& 29.4      	& 62.0   		& 0.952	\\
SAM-SS\cite{2019arXiv190900121C}           	& 43.8      	& 51.4    		& 28.9      	& 62.4   		& 0.970	\\
\midrule
VNS-GRU				                       	& 45.3 			& 50.3			& 29.2			& 62.6			& 0.977				\\
VNS-GRU$_r$			                       	& 45.4			& 51.7			& 29.3			& 62.9			& 0.986				\\
VNS-GRU$^{e}_r$                       		& \textbf{45.6}	& 51.8			& 29.7			& 63.2			& 0.992				\\
VNS-GRU$^{er}_r$                       		& 45.3			& \textbf{53.0}	& \textbf{29.9}	& \textbf{63.4}	&\textbf{0.998}	\\
\bottomrule
\end{tabular}
\end{center}
\end{table}

\subsubsection{Comparison on MSR-VTT}
The first three models in Table \ref{table:performanceMSRVTT} are the top-3 winners in MSR-VTT 2017 competition. 
HACA \cite{DBLP:conf/naacl/WangWW18} is a model with multi-level aligned multi-modal attention framework. 
TAMoE \cite{DBLP:conf/aaai/WangWZ0W19} learns to embed multiple topic-based experts into the model and implicitly transfers knowledge in seen activities to unknown ones. 

Our model VNS-GRU$^{er}_r$ also outperforms all the previous models on four metrics: BLEU-4, CIDEr, METEOR and ROUGE-L, 
and achieves gains over the closest rival SAM-SS by 3.4\%, 3.1\%, 3.5\% and 1.6\%, respectively. 
Note that our model surpasses CIDEnt-RL, which is directly optimized on CIDEr-related reward, on CIDEr.

\section{Conclusion}
In this work, we propose three methods to improve the decoder of video captioning model. 
The first is to embed variational dropout and layer normalization in RNN unit to prevent overfitting and sustain convergence speed. 
The second is an online method to select the best model for testing with comprehensive consideration on kinds of metrics. 
The last is a novel training scheme called \textit{professional learning}. 
In its first phase, all the training annotations for each video is equally treated in the process of optimization. 
In its second phase, the training algorithm aims to strengthen the strong points of the model, 
in other words, the samples with lower loss values have higher weights in optimization. 
Our model achieves state-of-the-art results on the popular video captioning benchmarks with a rich vocabulary and diversified sentences. 
However, in theory, \textit{profession learning} can be applied with other training algorithms together 
and it may further improve the performance of a model which we leave it for future research.

\ack 
We thank Hallbjorn Thor Gudmunsson for inspiration and extensive discussion. 
We also pay gratitude to the anonymous reviewers for their helpful evaluations.
This work was supported by the National Natural Science Foundation of China under Grant Nos. U19B2034, 61620106010,
61836014 and a grant from Samsung Research China, Beijing. 
\bibliography{193_paper}

\begin{thebibliography}{10}

\bibitem{DBLP:conf/cvpr/AafaqALGM19}
Nayyer Aafaq, Naveed Akhtar, Wei Liu, Syed~Zulqarnain Gilani, and Ajmal Mian,
  `Spatio-temporal dynamics and semantic attribute enriched visual encoding for
  video captioning', in {\em {CVPR}}, pp. 12487--12496, (2019).

\bibitem{Ba2016LayerN}
Jimmy Ba, Jamie~Ryan Kiros, and Geoffrey~E. Hinton, `Layer normalization', {\em
  ArXiv}, {\bf abs/1607.06450}, (2016).

\bibitem{DBLP:journals/corr/BaMK14}
Jimmy Ba, Volodymyr Mnih, and Koray Kavukcuoglu, `Multiple object recognition
  with visual attention', in {\em {ICLR}}, (2015).

\bibitem{DBLP:journals/corr/BahdanauCB14}
Dzmitry Bahdanau, Kyunghyun Cho, and Yoshua Bengio, `Neural machine translation
  by jointly learning to align and translate', in {\em {ICLR}}, (2015).

\bibitem{DBLP:conf/acl/BanerjeeL05}
Satanjeev Banerjee and Alon Lavie, `{METEOR:} an automatic metric for {MT}
  evaluation with improved correlation with human judgments', in {\em
  Proceedings of the Workshop on Intrinsic and Extrinsic Evaluation Measures
  for Machine Translation and/or Summarization@ACL}, pp. 65--72, (2005).

\bibitem{DBLP:conf/nips/BengioVJS15}
Samy Bengio, Oriol Vinyals, Navdeep Jaitly, and Noam Shazeer, `Scheduled
  sampling for sequence prediction with recurrent neural networks', in {\em
  NeurIPS}, pp. 1171--1179, (2015).

\bibitem{Bengio:2009:CL:1553374.1553380}
Yoshua Bengio, J{\'e}r\^{o}me Louradour, Ronan Collobert, and Jason Weston,
  `Curriculum learning', in {\em Proceedings of the 26th Annual International
  Conference on Machine Learning}, ICML '09, pp. 41--48, New York, NY, USA,
  (2009). ACM.

\bibitem{2019arXiv190900121C}
Haoran {Chen}, Ke~{Lin}, Alexander {Maye}, Jianming {Li}, and Xiaolin {Hu}, `{A
  Semantics-Assisted Video Captioning Model Trained with Scheduled Sampling}',
  {\em arXiv e-prints},  arXiv:1909.00121, (Aug 2019).

\bibitem{cho2014on}
Kyunghyun Cho, Bart Van~Merrienboer, Dzmitry Bahdanau, and Yoshua Bengio, `On
  the properties of neural machine translation: Encoder-decoder approaches',
  {\em arXiv: Computation and Language}, (2014).

\bibitem{DBLP:conf/emnlp/ChoMGBBSB14}
Kyunghyun Cho, Bart van Merrienboer, {\c{C}}aglar G{\"{u}}l{\c{c}}ehre, Dzmitry
  Bahdanau, Fethi Bougares, Holger Schwenk, and Yoshua Bengio, `Learning phrase
  representations using {RNN} encoder-decoder for statistical machine
  translation', in {\em {EMNLP}}, pp. 1724--1734, (2014).

\bibitem{chung2014empirical}
Junyoung Chung, Caglar Gulcehre, Kyunghyun Cho, and Yoshua Bengio, `Empirical
  evaluation of gated recurrent neural networks on sequence modeling', {\em
  arXiv: Neural and Evolutionary Computing}, (2014).

\bibitem{DBLP:conf/icml/GalG16}
Yarin Gal and Zoubin Ghahramani, `Dropout as a bayesian approximation:
  Representing model uncertainty in deep learning', in {\em {ICML}}, pp.
  1050--1059, (2016).

\bibitem{DBLP:conf/nips/GalG16}
Yarin Gal and Zoubin Ghahramani, `A theoretically grounded application of
  dropout in recurrent neural networks', in {\em NeurIPS}, pp. 1019--1027,
  (2016).

\bibitem{DBLP:conf/cvpr/GanGHPTGCD17}
Zhe Gan, Chuang Gan, Xiaodong He, Yunchen Pu, Kenneth Tran, Jianfeng Gao,
  Lawrence Carin, and Li~Deng, `Semantic compositional networks for visual
  captioning', in {\em {CVPR}}, pp. 1141--1150, (2017).

\bibitem{DBLP:journals/tmm/GaoGZXS17}
Lianli Gao, Zhao Guo, Hanwang Zhang, Xing Xu, and Heng~Tao Shen, `Video
  captioning with attention-based {LSTM} and semantic consistency', {\em {IEEE}
  Trans. Multimedia}, {\bf 19}(9),  2045--2055, (2017).

\bibitem{DBLP:conf/nips/GoyalLZZCB16}
Anirudh Goyal, Alex Lamb, Ying Zhang, Saizheng Zhang, Aaron~C. Courville, and
  Yoshua Bengio, `Professor forcing: {A} new algorithm for training recurrent
  networks', in {\em NeurIPS}, pp. 4601--4609, (2016).

\bibitem{DBLP:conf/iccv/GuadarramaKMVMDS13}
Sergio Guadarrama, Niveda Krishnamoorthy, Girish Malkarnenkar, Subhashini
  Venugopalan, Raymond~J. Mooney, Trevor Darrell, and Kate Saenko,
  `Youtube2text: Recognizing and describing arbitrary activities using semantic
  hierarchies and zero-shot recognition', in {\em {ICCV}}, pp. 2712--2719,
  (2013).

\bibitem{DBLP:conf/cvpr/HeZRS16}
Kaiming He, Xiangyu Zhang, Shaoqing Ren, and Jian Sun, `Deep residual learning
  for image recognition', in {\em {CVPR}}, pp. 770--778, (2016).

\bibitem{DBLP:journals/neco/HochreiterS97}
Sepp Hochreiter and J{\"{u}}rgen Schmidhuber, `Long short-term memory', {\em
  Neural Computation}, {\bf 9}(8),  1735--1780, (1997).

\bibitem{DBLP:conf/iccv/HoriHLZHHMS17}
Chiori Hori, Takaaki Hori, Teng{-}Yok Lee, Ziming Zhang, Bret Harsham, John~R.
  Hershey, Tim~K. Marks, and Kazuhiko Sumi, `Attention-based multimodal fusion
  for video description', in {\em {ICCV}}, pp. 4203--4212, (2017).

\bibitem{DBLP:journals/corr/Huszar15}
Ferenc Huszar, `How (not) to train your generative model: Scheduled sampling,
  likelihood, adversary?', {\em CoRR}, {\bf abs/1511.05101}, (2015).

\bibitem{DBLP:conf/icml/IoffeS15}
Sergey Ioffe and Christian Szegedy, `Batch normalization: Accelerating deep
  network training by reducing internal covariate shift', in {\em {ICML}}, pp.
  448--456, (2015).

\bibitem{lin-2004-rouge}
Chin-Yew Lin, `{ROUGE}: A package for automatic evaluation of summaries', in
  {\em Text Summarization Branches Out}, pp. 74--81, Barcelona, Spain, (July
  2004). Association for Computational Linguistics.

\bibitem{DBLP:conf/cvpr/MaKMKAG18}
Chih{-}Yao Ma, Asim Kadav, Iain Melvin, Zsolt Kira, Ghassan AlRegib, and
  Hans~Peter Graf, `Attend and interact: Higher-order object interactions for
  video understanding', in {\em {CVPR}}, pp. 6790--6800, (2018).

\bibitem{DBLP:conf/asru/MoonCLS15}
Taesup Moon, Heeyoul Choi, Hoshik Lee, and Inchul Song, `{RNNDROP:} {A} novel
  dropout for {RNNS} in {ASR}', in {\em {IEEE} Workshop on Automatic Speech
  Recognition and Understanding}, pp. 65--70, (2015).

\bibitem{DBLP:conf/acl/PapineniRWZ02}
Kishore Papineni, Salim Roukos, Todd Ward, and Wei{-}Jing Zhu, `Bleu: a method
  for automatic evaluation of machine translation', in {\em ACL}, pp. 311--318,
  (2002).

\bibitem{DBLP:conf/acl/PasunuruB17}
Ramakanth Pasunuru and Mohit Bansal, `Multi-task video captioning with video
  and entailment generation', in {\em {ACL}}, pp. 1273--1283, (2017).

\bibitem{DBLP:conf/emnlp/PasunuruB17}
Ramakanth Pasunuru and Mohit Bansal, `Reinforced video captioning with
  entailment rewards', in {\em {EMNLP}}, pp. 979--985, (2017).

\bibitem{DBLP:conf/cvpr/PeiZWKST19}
Wenjie Pei, Jiyuan Zhang, Xiangrong Wang, Lei Ke, Xiaoyong Shen, and Yu{-}Wing
  Tai, `Memory-attended recurrent network for video captioning', in {\em
  {CVPR}}, pp. 8347--8356, (2019).

\bibitem{DBLP:conf/cvpr/RennieMMRG17}
Steven~J. Rennie, Etienne Marcheret, Youssef Mroueh, Jerret Ross, and Vaibhava
  Goel, `Self-critical sequence training for image captioning', in {\em
  {CVPR}}, pp. 1179--1195, (2017).

\bibitem{DBLP:conf/coling/SemeniutaSB16}
Stanislau Semeniuta, Aliaksei Severyn, and Erhardt Barth, `Recurrent dropout
  without memory loss', in {\em {COLING}}, pp. 1757--1766, (2016).

\bibitem{DBLP:journals/jmlr/SrivastavaHKSS14}
Nitish Srivastava, Geoffrey~E. Hinton, Alex Krizhevsky, Ilya Sutskever, and
  Ruslan Salakhutdinov, `Dropout: a simple way to prevent neural networks from
  overfitting', {\em J. Mach. Learn. Res.}, {\bf 15}(1),  1929--1958, (2014).

\bibitem{DBLP:conf/cvpr/SzegedyLJSRAEVR15}
Christian Szegedy, Wei Liu, Yangqing Jia, Pierre Sermanet, Scott~E. Reed,
  Dragomir Anguelov, Dumitru Erhan, Vincent Vanhoucke, and Andrew Rabinovich,
  `Going deeper with convolutions', in {\em {CVPR}}, pp. 1--9, (2015).

\bibitem{DBLP:conf/cvpr/VedantamZP15}
Ramakrishna Vedantam, C.~Lawrence Zitnick, and Devi Parikh, `Cider:
  Consensus-based image description evaluation', in {\em {CVPR}}, pp.
  4566--4575, (2015).

\bibitem{venugopalan-etal-2015-translating}
Subhashini Venugopalan, Huijuan Xu, Jeff Donahue, Marcus Rohrbach, Raymond
  Mooney, and Kate Saenko, `Translating videos to natural language using deep
  recurrent neural networks', in {\em NAACL}, pp. 1494--1504, Denver, Colorado,
  (May{--}June 2015). Association for Computational Linguistics.

\bibitem{DBLP:conf/cvpr/WangCWWW18}
Xin Wang, Wenhu Chen, Jiawei Wu, Yuan{-}Fang Wang, and William~Yang Wang,
  `Video captioning via hierarchical reinforcement learning', in {\em {CVPR}},
  pp. 4213--4222, (2018).

\bibitem{DBLP:conf/naacl/WangWW18}
Xin Wang, Yuan{-}Fang Wang, and William~Yang Wang, `Watch, listen, and
  describe: Globally and locally aligned cross-modal attentions for video
  captioning', in {\em NAACL-HLT}, pp. 795--801, (2018).

\bibitem{DBLP:conf/aaai/WangWZ0W19}
Xin Wang, Jiawei Wu, Da~Zhang, Yu~Su, and William~Yang Wang, `Learning to
  compose topic-aware mixture of experts for zero-shot video captioning', in
  {\em {AAAI}}, pp. 8965--8972, (2019).

\bibitem{DBLP:journals/neco/WilliamsZ89}
Ronald~J. Williams and David Zipser, `A learning algorithm for continually
  running fully recurrent neural networks', {\em Neural Computation}, {\bf
  1}(2),  270--280, (1989).

\bibitem{DBLP:conf/acl/WuRLS19}
Chen Wu, Xuancheng Ren, Fuli Luo, and Xu~Sun, `A hierarchical reinforced
  sequence operation method for unsupervised text style transfer', in {\em
  {ACL}}, pp. 4873--4883, (2019).

\bibitem{DBLP:journals/ijon/WuWCSSW18}
Chunlei Wu, Yiwei Wei, Xiaoliang Chu, Weichen Sun, Fei Su, and Leiquan Wang,
  `Hierarchical attention-based multimodal fusion for video captioning', {\em
  Neurocomputing}, {\bf 315},  362--370, (2018).

\bibitem{DBLP:conf/cvpr/XieGDTH17}
Saining Xie, Ross~B. Girshick, Piotr Doll{\'{a}}r, Zhuowen Tu, and Kaiming He,
  `Aggregated residual transformations for deep neural networks', in {\em
  {CVPR}}, pp. 5987--5995, (2017).

\bibitem{DBLP:conf/cvpr/XuMYR16}
Jun Xu, Tao Mei, Ting Yao, and Yong Rui, `{MSR-VTT:} {A} large video
  description dataset for bridging video and language', in {\em {CVPR}}, pp.
  5288--5296, (2016).

\bibitem{DBLP:conf/icml/XuBKCCSZB15}
Kelvin Xu, Jimmy Ba, Ryan Kiros, Kyunghyun Cho, Aaron~C. Courville, Ruslan
  Salakhutdinov, Richard~S. Zemel, and Yoshua Bengio, `Show, attend and tell:
  Neural image caption generation with visual attention', in {\em {ICML}}, pp.
  2048--2057, (2015).

\bibitem{DBLP:conf/iccv/YaoTCBPLC15}
Li~Yao, Atousa Torabi, Kyunghyun Cho, Nicolas Ballas, Christopher~J. Pal, Hugo
  Larochelle, and Aaron~C. Courville, `Describing videos by exploiting temporal
  structure', in {\em {ICCV}}, pp. 4507--4515, (2015).

\bibitem{DBLP:journals/corr/ZarembaSV14}
Wojciech Zaremba, Ilya Sutskever, and Oriol Vinyals, `Recurrent neural network
  regularization', {\em CoRR}, {\bf abs/1409.2329}, (2014).

\bibitem{DBLP:conf/eccv/ZolfaghariSB18}
Mohammadreza Zolfaghari, Kamaljeet Singh, and Thomas Brox, `{ECO:} efficient
  convolutional network for online video understanding', in {\em {ECCV}}, pp.
  713--730, (2018).

\end{thebibliography}
\end{document}